\documentclass{article}

     \usepackage[preprint]{neurips_2019}

\usepackage{multirow}
\usepackage[utf8]{inputenc} 
\usepackage[T1]{fontenc}    
\usepackage{hyperref}       
\usepackage{url}            
\usepackage{booktabs}       
\usepackage{amsfonts}       
\usepackage{nicefrac}       
\usepackage{microtype} 
\usepackage{graphicx}
\newcommand{\comment}[1]{}
\title{A Retrieval-based Approach to Goal-Oriented Dialogue}
\title{Retrieval-based Goal-Oriented Dialogue Generation}
\author{%
  Ana V. Gonzalez, Isabelle Augenstein and Anders Søgaard \\
  Department of Computer Science, University of Copenhagen\\
  \texttt{\{ana,augenstein,soegaard\}@di.ku.dk} \\
}

\begin{document}

\maketitle

\begin{abstract}
  Most research on dialogue has focused either on dialogue generation for open-ended chit chat or on state tracking for goal-directed dialogue. In this work, we explore a hybrid approach to goal-oriented dialogue {\em generation} that combines retrieval from past history with a hierarchical, neural encoder-decoder architecture. 
 We evaluate this approach in the customer support domain using the Multiwoz dataset \citep{budzianowski2018multiwoz}.  
 We show that adding this retrieval step to a hierarchical, neural encoder-decoder architecture leads to significant improvements, including responses that are rated more appropriate and fluent by human evaluators. Finally, we compare our retrieval-based model to various semantically conditioned models explicitly using past dialog act information, and find that our proposed model is competitive with the current state of the art \citep{chen2019semantically}, while not requiring explicit labels about past machine acts.
\end{abstract}

\section{Introduction}

Dialogue systems have become a very popular research topic in recent years with the rapid improvement of personal assistants and the growing demand for online customer support. However, research has been split in two subfields \citep{chen2017survey}: models presented for generation of open-ended conversations \citep{serban2015hierarchical, li2017adversarial, shibata2009dialog,sugiyama2013open, ritter2011data} and work on solving goal-oriented dialogue through dialogue management pipelines that include dialogue state tracking and dialogue policy \citep{ren2018towards,mrkvsic2016neural, yoshino2016dialogue,henderson2013deep, sun2014generalized, zhao2016, ren2013dialog}. 

Dialogue state tracking has often been limited to detection of user intention, as well as learning a dialogue policy to determine what actions the system should take based on the detected user intent. 
Dialogue generation for open ended conversation, in contrast, has largely relied on transduction architectures originally developed for machine translation (MT) \citep{shang2015responding,zhang2018coherence,wen2018kb}. Such architectures offer flexibility because of their ability to encode an utterance into a fixed-sized vector representation, and decoding it into a variable length sequence that is linguistically very different from the input utterance. However, MT-based approaches often lack the ability to encode the context in which the current utterance occurs. This can lead to repetitive and meaningless responses \citep{li2015hierarchical,Lowe:ea:17,wen2018kb}.

 This observation has led researchers to extend simple encoder-decoder models to include context in order to deal with generation of larger structured texts such as paragraphs and documents \citep{li2015hierarchical,serban2016building,serban2017hierarchical}. Many of these models work by encoding information at multiple levels, i.e., using both a context encoder and a last-utterance encoder, passing both encodings to a decoder that predicts the next turn. Such hierarchical methods have proven to be useful for open-ended chit chat, but were not designed for goal-oriented dialogue, where responses need not only be coherent, but also relevant. 
 
 In \textit{goal-directed dialogue generation}, there is often one (context-dependent) right answer to a question (e.g., {\em How many types of insurance do you offer?}); in chit-chat, there are many good answers to questions (e.g., {\em What do you want to talk about today?}). We therefore hypothesize that in personal assistants and customer support, it is beneficial to increase the inductive bias of the dialogue generation model and its dependency on past conversations in order to keep responses relevant. We do so by designing a novel, hybrid dialogue generation model that conditions decoding on retrieved examplars from past history. 
 
 Although retrieval approaches to dialogue generation have been introduced before, they have typically been used to add more variety to the kind of answers the model can generate in open ended conversations \citep{ritter2011data, weston2018retrieve}. Our model, in contrast, is designed for the purpose of goal oriented dialogue in the customer support domain. It is a hierarchical neural model with an information retrieval component that \textit{retrieves the most informative prior turns and conditions on those}. This simple approach increases inductive bias and alleviates problems arising from long-context dependencies. 
 
 We show that an information retrieval step leads to improvements over traditional dialogue generation models intended for open ended chit chat when evaluated on BLEU and different embedding metrics. In addition, we evaluate our generated responses based on the Request/Inform success rate typically used in dialogue state tracking, and again, we show performance close to the more complex state-of-the-art model which contrary to our proposed model, makes use of annotated labels. Finally, based on our human evaluations, we show that a simple retrieval step leads to system responses that are more fluent and appropriate than the responses generated by the hierarchical encoder-decoder model.
 
\begin{figure}
\begin{center}
\includegraphics[width=\textwidth]{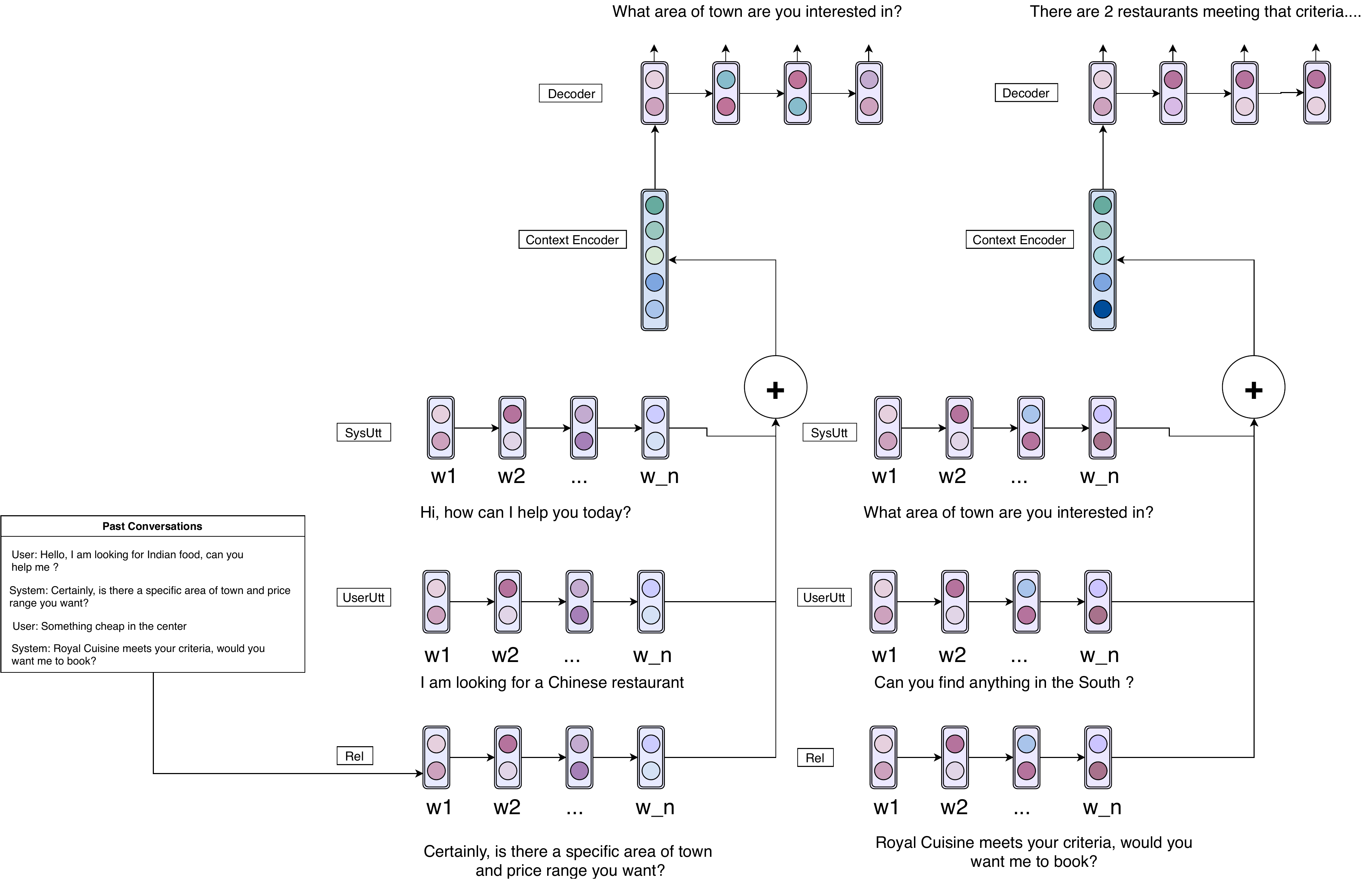}
\end{center}
\caption{Our model is similar to HRED \citep{sordoni2015hierarchical}, we include an utterance encoder, a context encoder and a decoder, however, unlike HRED, our model include a simple, yet effective retrieval step used to condition the decoder to generate responses that are more appropriate for a specific domain and context.}
\end{figure}

\section{Model Description}
We want to adapt commonly used models for  dialogue generation to the task of goal-oriented dialogue by providing a simple way of integrating past examples in a context-aware dialogue generation model. We extend the Hierarchical Recurrent Encoder-Decoder (HRED) model presented by \citet{sordoni2015hierarchical} for query suggestion,  subsequently adopted for dialogue by \citet{serban2016building}, which has shown to be a strong baseline for the task of dialogue generation. In line with previous research, we consider a dialogue $D$ between two speakers composed of $n$ utterances so that $D = [U_1, ... , U_n]$ and each utterance $U_i$ composed of $k_i$ tokens so that $U_n = [t_{i,1}, t_{i, 2}, ... , t_{i,k_i}]$. Each token $t_{i, k_i}$ represents a word from a set vocabulary. Preliminary experiments showed that limiting contexts to three utterances gave the best results, therefore all results presented use $n=3$. This is also in line with most previous work on dialogue generation and classification \citep{serban2016building}.

\subsection{HRED}
HRED \citep{sordoni2015hierarchical} consists of an encoder RNN, which maps an utterance to a vector, a context RNN, which summarizes the dialogue history by keeping track of the previous hidden states, and a decoder RNN, which decodes the hidden state of the context RNN. Given a dialogue consisting of three utterances -- a system response, a user response, and a second system response, $\langle s_1,u,s_2\rangle$ -- 
the goal is to predict the system utterance $s_2$ given the context $\langle s_1,u\rangle$. 
The utterance encoder takes in a single utterance and outputs a vector representation . The representations of each utterance are concatenated so that we end up with an array that is then fed to the context encoder. The context encoder outputs a global context, which is then fed into the decoder.   Just as in previous work \citep{serban2017hierarchical, serban2016building, serban2015hierarchical, shen2017conditional}, we use GRUs \citep{cholearning} for our encoder, context and decoder RNNs. All modules share parameters.

\subsection{Exemplar-HRED}
In this study, we want to enhance HRED with a simple yet efficient information retrieval step. As already mentioned, similar approaches have been presented with the goal of incorporating factual information into open-ended conversations \citep{weston2018retrieve}, to add variety and more topics to the conversation. 

We hypothesize that using exemplar information is also beneficial for multi-domain goal-oriented systems. More specifically, we want to be able to inform our generation model about previous responses to similar utterances, biasing it towards past responses.  For each user utterance, we extract the ten most similar past user utterances from the training set using approximate nearest neighbor search  \citep{Indyk:1998:ANN:276698.276876}. 
We approximate a point $p \in S$ by specifying some error margin $\epsilon > 0 $ so that  $ dist(p,q) \leq (1+\epsilon)(dist(p*, q)) $, 
where $p*$ is the real nearest neighbor. 
Because we use approximate search, we rerank the retrieved utterances using a feed-forward ranking model, introduced in \citet{gonzalez2018strong}. Their ranking model is a multi-task model, which relies on simple textual similarity measures combined in a multi-layered perceptron architecture. The model nevertheless achieves state-of-the-art performance on question relevancy ranking. In the end, we take the top user utterance, and return its response as the example to be used in our model. 

For goal-oriented dialogue generation, our proposed model uses the same architecture as the HRED baseline, however, we include an additional RNN, which encodes the top example response. We feed the representation of the example RNN context into the context RNN and feed this representation into the decoder. Just as in the baseline model, the utterance encoder outputs a vector representation. Additionally, we encode the exemplar into a vector using the example encoder.  The representations of each utterance are concatenated so that we end up with an array that includes dialogue context and exemplar information, all of which is then fed to the context encoder. The global context is then fed into the decoder.

For all experiments, we use the MultiWoz dataset for goal oriented dialogue \citep{budzianowski2018multiwoz}, which we describe in more detail in the next section. 
Our model uses the Adam optimizer \citep{KingmaB14} for all encoders. All our encoders are one layer RNNs. In addition, we use a dropout rate of 0.3, and a learning rate of 0.001. We set a maximum of 50 epochs, however, we use early stopping with a patience of 10. Most of our models converge by epoch 30. We use greedy search to generate the response during testing. More implementation details as well as our predicted utterances for each system can be found in the link provided \footnote{\url{https://github.com/anavaleriagonzalez/exemplar_dialog}}


\section{Experiments}

\paragraph{Dataset}
We use the MultiWoz dialogue corpus \citep{budzianowski2018multiwoz}, which consists of 10,438 dialogues  spanning several domains and annotated with dialogue states and acts. We train on 8,438 dialogues, and use 1000 dialogues for development and 1000 dialogues for testing. Although the data is primarily intended for dialogue state tracking and learning a dialogue policy, \citet{budzianowski2018multiwoz} also mention its potential as a benchmark for end-to-end dialogue due to the fact that it contains about 115k turns in total, which is larger than many \textit{structured} dialogue corpora available. This makes it a good choice for hybrid approaches in generation and goal oriented dialogue. The MultiWOZ dataset is also a much more difficult dataset than the current benchmarks for goal oriented dialogue, as it spans about 7 different customer support domains and conversations are not limited to a single domain. In line with previous work in goal oriented dialog and recent work using this dataset, we delexicalize the utterances to remove phone numbers, reference numbers and train ids. As opposed to other studies, we only delexicalize these three slots since these were significantly increasing the size of the vocabulary. For delexicalizing, we use the ontology provided with the data and replace the value with the slot names using regular expressions. We do not delexicalize times, prices, postcodes and distinct names of restaurants and hotels. This also makes our generation task more difficult.

\paragraph{Baselines}
In this study, we are interested in simple ways of providing our dialogue generation model with enough inductive bias in order to generate fluent and on-topic responses for goal oriented dialogue. Encoder-decoder architectures work well when it comes to providing generic fluent responses for open ended conversations, however, in goal-oriented dialogue, it is also necessary for the system to remain on topic. Additionally, when training a single model on different domains, this becomes more difficult. The original HRED model \citep{sordoni2015hierarchical,serban2017hierarchical} adapted for dialogue performs well when it comes to open ended conversations as well as when trained on very large corpora (millions of utterances) \citep{lowe2015ubuntu,lison2016opensubtitles2016}. In our setup however, we train the HRED model using a smaller dataset containing goal oriented dialogues in 7 different domains. We compare this model with our proposed exemplar-based model. In addition, for the BLEU metric we include the results of a transformer model\citep{vaswani2017attention} that uses dialogue context to condition the decoder as well as a LSTM that uses dialogue context and incorporates belief state and KB results as additional inputs \citep{budzianowski2018multiwoz}.

\begin{table*}[h!]
\begin{center}
\begin{tabular}{|p{5.5cm}|p{2cm}|p{2cm}|}
\hline

\textbf{Metric} &\textbf{HRED} & \textbf{Exemplar-HRED}\\
\hline
BLEU & 23.6 &\textbf{24.1} \\

Vector Extrema & .59 &  \textbf{.65}\\

Average Embedding Similarity &  .93 & \textbf{.95}\\

Greedy Matching &  23.1 & \textbf{23.9} \\

Human Eval- Fluency & 0.19& \textbf{0.58}\\

Human Eval- Appropriateness & 0.14&\textbf{0.59} \\

\hline

\end{tabular}
\vspace{.5cm}
\caption{The results of our dialogue generation experiments comparing HRED \protect{\citet{sordoni2015hierarchical,serban2016building}} to our proposed exemplar-based model. We present resultd for standard metrics used in dialogue generation. For all the metrics we observe improvements over the strong baseline, with our best improvement of 6 percent in the vector extrema metric}
\label{tab:results}
\end{center}
\end{table*}

\section{Results} Overall, we found that in most cases, our simple model leads to significant improvements over the standard metrics \citep{liu2016not}; see Table~1 for the results. Although we are tackling goal-oriented dialogue, traditional metrics for goal oriented dialogue rely on human-generated supervision i.e. slot-value pair labels or dialogue act labels. Word overlap metrics such as the ones used for machine translation are often used to evaluate the quality of dialogue generation, however, these standard metrics tend to have very weak correlation with human judgment. In any case, we include some of these, as well as word embedding metrics for comparison. For the standard metrics, we use the evaluation scripts from \citep{serban2016building} \footnote{https://github.com/julianser/hed-dlg-truncated/tree/master/Evaluation}. We observe that the retrieval model is consistently better across all scenarios and metrics. In addition to these metrics, we assess our performance using the dialogue success metrics typically used in belief tracking \citep{budzianowski2018multiwoz}. We briefly explain these metrics further.

\paragraph{BLEU}
BLEU \citep{Papineni02bleu:a} is typically used for machine translation and has subsequently been used to evaluate the performance of many dialogue generation systems \citep{galley2015deltableu,serban2017hierarchical, serban2016building}. BLEU analyzes co-occurrences of n-grams in a reference sequence and a hypothesis. For all datasets, we see improvements with BLEU. It uses a modified precision to account for the differences in length between reference and generated output. Given a reference sentence $s$ and a hypothesis sentence $\hat{s}$, we can denote the n-gram precision $P_{n}(s, \hat{s})$ as:
$$P_{n}(s, \hat{s}) =  \frac{\sum_{q}min(h(q, s),(h(q, \hat{s}))}{\sum_{q} h(q, s)} $$
where q is the index of all possible n-grams, and h(q,s) is the number of n-grams in s. 

\paragraph{Average Word Embedding Similarity}
We follow \citet{liu2016not} and obtain the average embedding $e_s$ for the reference sentence $s$ by averaging the word embeddings $e_w$ for each token $w$ in $s$. We do the same for the predicted output $\hat{s}$ and obtain the final similarity score by computing cosine similarity of the two resulting vectors. Again, Exemplar-HRED is consistently superior yielding almost a 2 percent improvement over the best baseline model. 

\paragraph{Vector Extrema}

We also compute the cosine similarity between the vector extrema of the reference and the hypothesis, again following \citet{liu2016not}. The goal of this metric as described by the authors is to consider informative words rather than common words, since the vectors for common words will tend to be pulled towards the zero vector. Our exemplar model achieves the largest improvement for this metric, with a gain of 6 percent over the baseline model.

\paragraph{Greedy Matching}
In greedy matching \citep{liu2016not}, given two sequences $s$ and $\hat{s}$, each token $w \in s$ is matched with each token  $\hat{w} \in \hat{s}$ by computing the cosine similarity of the corresponding word embeddings $emb_w$ and $emb_{\hat{w}}$. The local match $g(s, \hat{s})$ is the word embedding with the maximum cosine similarity. We compute in both directions and the total score is:
$$ G(s, \hat{s}) =\frac{g(s, \hat{s}) +g(\hat{s}, s)}{2} $$
This metric is used to favour key words. Our best model shows only small improvements on this metric.

\paragraph{Human Evaluation}
In addition to the previously mentioned standard metrics, we also evaluate the performance of our baseline and the exemplar-based models using human evaluations. We extract 100 baseline and exemplar model system responses at random. We ask the $7$ evaluators to 1) pick the response that is more fluent and grammatically correct and 2) pick the response that achieves the goal given the context of the conversation. We provide the context of System and User utterances, and ask the evaluators to pick one of 4 options: 1) the output of the baseline, 2) the output of the exemplar model, 3) both, 4) none. The order of the options was shuffled.

Overall, we found that when it came to fluency, the evaluators perceived that 58\% of the time, the exemplar response was better. The baseline beat the exemplar based response for 19 percent of the evaluated dialogs and the rest of the dialogs either both or none were picked. For appropriateness we see a similar pattern. Evaluators perceived the response produced by the exemplar model as the more appropriate one given the context, for 59 percent of the evaluated dialogs. The baseline beat the proposed model only 14 percent of the time. These results can also be found on table \ref{tab:results}

\paragraph{Dialogue success: inform/request}
Traditional goal-oriented dialogue systems based on prediction of slots and dialogue acts are typically evaluated on the accuracy of predicting these as labels, as well as their success at the end of the dialogue. Dialogue success is measured by how many correct inform/request slots a model can generate in a conversation in comparison to the ground truth. An inform slot is one that provides the user with a specific item for example the inform slots "food" and "area" i.e. (food=“Chinese”, area="center”) informs the user that there is a Chinese restaurant in the center. On the other hand, a request slot is a slot that specifies what information is needed for the system to achieve the user goal. For example, for booking a train, the system needs to know the departure location and the destination. The slots "departure" and "destination" would be the request slots in this case. 

For goal-oriented generation, many of the models evaluated using the Inform/Request metrics have made use of structured data to semantically condition the generation model in order to generate better responses \citep{wen2015semantically,chen2019semantically,budzianowski2018multiwoz}. \cite{wen2015semantically} proposed to encode each individual dialog act as a unique vector and use it as an extra input feature, in order to influence the generated response. This method has been shown to work well when tested on single domains where the label space is limited to a few dialog acts. However, as the label space grows, using a one-hot encoding representation of the dialog act is not scalable. To deal with this problem, \cite{chen2019semantically} introduced a semantically conditioned generation model using Hierarchical Disentangled Self-Attention (HDSA) . This model deals with the large label space by representing dialog acts using a multi-layer hierarchical graph that merges cross-branch nodes. For example, the distinct trees for {\sc hotel-recommend-area} 
and
{\sc attraction-recommend-area} can be merged at the second and third levels sharing semantic information about actions and slots but maintaining specific information about the domains separate. This information can then be used as part of an attention mechanism when generating a response. This model achieves the state-of-the-art result for generation in both BLEU and Inform/Request metrics. 
 
As we are concerned with improving dialogue generation for goal oriented dialogue, we are interested in assessing how our simple approach compares to models explicitly using dialog acts as extra information. We compute the inform/request accuracy and compare to the state-of-the-art \citep{chen2019semantically} as well as other baseline models. \cite{chen2019semantically} present experiments conditioning both on predicted acts as well as ground truth past acts. We include both of these as well as the performance of our baseline and proposed model in table \ref{tab:results-success}. We divide the results into models using act information to condition the language generation and models that do not.
 
\begin{table}[h!]
\begin{center}
\begin{tabular}{|p{1cm}|p{6cm}|p{1cm}|p{1cm}|p{1cm}|}
\hline
& Model & Inform & Request & BLEU \\

\hline
\multirow{ 3}{*}{No act} & 3-layer Transformer (Vaswani et al., 2017) & 71.1 & 59.9 & 19.1 \\
&HRED &60.4 & 44.5 &23.6  \\
&Exemplar-HRED &77.6 & 70.1 &24.1  \\

\hline
\multirow{ 4}{*}{Act} &LSTM (Budzianowski et al., 2018) & 71.2 & 60.2 & 18.8 \\
& SC-LSTM (Wen et al., 2015) & 74.5 &62.5 & 20.5  \\
&HDSA-predicted & 82.9 & 68.9 & 23.6  \\
&HDSA-groundtruth & 87.9 & 78.0 & 30.4  \\

\hline

\end{tabular}
\vspace{.5cm}
\caption{Inform/request results divided into two section. The top models do not make use of any past dialog acts to condition the decoder to generate a response. The models at the bottom use dialog acts, and belief state in order to generate better responses}
\label{tab:results-success}
\end{center}
\end{table}

\section{Discussion}
 As shown in table \ref{tab:results}, our simplest proposed model achieved the largest improvements over the baseline when it came to the average embedding similarity and vector extrema similarity. As it is hard to interpret what the difference in performance of each model is based on standard dialogue metrics we examined the output to spot the major differences in response generation of our proposed models versus the baseline. We looked at the responses generated by our proposed models that had the highest score of these metrics and compared to the response generated by the baseline for that same dialogue. Overall we found that the baseline models tend to generate responses containing slots and values for the wrong domain. In addition, by examining the outputs we could see that the vector extrema metric is very sensitive when it comes to slight differences in the references and prediction. We found that this metric was more indicative of model performance than embedding similarity. We present some example outputs in table \ref{table:examples-extrema}. 

As mentioned earlier, from manual inspection of the outputs we observed that the the exemplar model is able to stay within the correct domain of the conversation and returns information within that domain that is more appropriate given the conversation context. This was confirmed by the human evaluations and also the Inform/Request metrics. When comparing the performance of the exemplar-based model to models that do not use information about past acts to condition the decoder, we observe that including a simple retrieval step leads to very large gains in the success of providing the inform/request slots.. 
In addition, the exemplar model performs better than \cite{budzianowski2018multiwoz}, which uses information of the belief state of the conversation as extra features. More interestingly, our proposed model performs better than the state-of-the-art when it comes to providing the request slots. It also outperforms this same model when evaluated on BLEU; however, it still falls behind the state-of-the-art when it comes to providing inform slots. Overall, we find that our model remains competitive without requiring turn labels.
\begin{table*}[h!]

\begin{center}

\begin{tabular}{p{4cm}| p{4cm}| p{4cm} }

\textbf{Gold Response} & \textbf{Exemplar-HRED} & \textbf{HRED} \\
\hline

 your booking was successful and your reference number is hotel-reference . is there anything else i can help you with ? 

&  your booking was successful . your reference number is hotel-reference . is there anything else i can help with ? 

& the booking was successful . the table will be reserved for 15 minutes . reference number is : restaurant-reference . \\

\hline
\textbf{Avg emb score} & 0.99 &  0.93 \\
\hline
\textbf{Vector Extrema} & 0.99 &  0.65 \\
\hline
\hline
booking was successful . the table will be reserved for 15 minutes . your reference number is : restaurant-reference .

&   your reservation was successful . the table will be reserved for 15 minutes . reference number is : restaurant-reference .

&   the train-trainid is train-trainid and arrives at stansted airport at the airport at 15:08 . would you like a booking \\
\hline
\textbf{Avg emb score} & 0.99 &  0.91 \\
\hline
\textbf{Vector Extrema} & 0.90 &  0.56 \\
\hline
\hline
there is no option for me to book it . is there anything else i can help you with ? &

certainly ! the reference number is restaurant-reference . can i help you with anything else ? &

the train-trainid leaves at 09:01 and arrives in cambridge by 09:52 . would that work ? \\
\hline
\textbf{Avg emb score} & 0.99 &  0.92 \\
\hline
\textbf{Vector Extrema} & 0.74 &  0.51 \\
\hline
\hline

yes , the address is saint andrew 's street and the phone number is hotel-phone . is there anything else i can assist you with today ? &

the address is saint andrew 's street , and the phone number is hotel-phone . can i help you with &

your taxi is booked for you . the contact number is taxi-phone . is there anything else i can help\\

\hline
\textbf{Avg emb score} &  0.99 &  0.98 \\
\hline
\textbf{Vector extrema} &  0.98 &  0.75 \\
\hline
\hline

the travel time is 88 minutes .  &

the travel time is 79 minutes . &

what time would you like to travel ?  \\
\hline
\textbf{Avg emb score} &  0.99 &  0.88 \\
\hline
\textbf{Vector Extrema} & 0.97 &  0.59 \\
\hline
\hline
where will you leave from ? &

sure , when would you like to leave ? &

have a nice stay . \\
\hline

\textbf{Avg emb score} &  0.98 &  0.95 \\
\hline
\textbf{Vector Extrema} & 0.94 &  0.74 \\
\hline
\hline
\end{tabular}

\caption{Examples of responses generated by both the baseline and our proposed model. By examining the outputs, it becomes noticeable that the baseline model tends to generate responses that are not precise about the current domain of the conversation (hotel, taxi booking, trains, restaurant, etc). The Exemplar-HRED model on the other hand becomes more accurate in responding for the correct domain. Here we present a few examples of responses given the same context, and provide their scores (similarity to reference) for comparison}
\label{table:examples-extrema}
\end{center}
\end{table*}

\section{Related Work}

Dialogue generation has relied on transduction architectures originally developed for machine translation (MT) \citep{shang2015responding,zhang2018coherence,wen2018kb}.  
Open domain dialogue systems aim to generate fluent and meaningful responses, however this has proven a challenging task. Most systems are able to generate coherent responses that are somewhat meaningless and at best entertaining \citep{Lowe:ea:17,wen2018kb,serban2016building}. Much of the research on dialogue generation has tried to tackle this problem by predicting an utterance based on some dialogue history \citep{vinyals2015neural,shang2015neural, luan2016lstm, serban2016building}. We extend such an architecture to also include past history, in order to avoid generating too generic responses. 

Most research on goal-oriented dialogue has focused almost exclusively on dialogue state tracking and dialogue policy learning \citep{sun2014generalized,sun2016hybrid, Li:ea:17,henderson2015machine,henderson2014word, rastogi2017scalable, mrkvsic2016neural, yoshino2016dialogue}. Dialogue state tracking consists of detecting the user intent and tends to rely on turn-level supervision and a preset number of possible slot and value pairs which limits the flexibility of such chatbots, including their ability to respond to informal chit chat, as well as transferring knowledge across domains. There has been some work in the past few years that has attempted to address these problems by introducing methods that focus on domain adaptation as well as introducing new data to make this task more accessible   \citep{rastogi2017scalable,rastogi2019scaling, budzianowski2018multiwoz, mrkvsic2015multi}. Recent approaches have also introduced methods for representing slot and value pairs that do not rely on a preset ontology \citep{ren2018towards, mrkvsic2017neural}, in an attempt to add flexibility. In our research, we acknowledge the importance of this added flexibility in goal oriented dialogue and propose a method for generating goal oriented responses without having turn level supervision.

The idea of combining text generation with past experience has been explored before. \citet{white1998exemplars} used a set of hand crafted examples in order to generate responses through templates. More recently, \citet{Song2016} also explored a hybrid system with an information retrieval component, but their system is very different: It uses a complex ranking system at a high computational cost, requires a post-reranking component to exploit previous dialogue turns (about half of which are copied over as predictions), and they only evaluate their system in a chit-chat set-up, reporting only BLEU scores. In a similar paper, \citep{weston2018retrieve} tried to move away from short generic answers in order to make a chit-chat generation model more entertaining by using an information retrieval component, to introduce relevant facts. In addition, a similar method was recently shown to improve other generation tasks such as summarization. In \cite{subramanian2019extractive}, the authors show that a simple extractive step introduces enough inductive bias for an abstractive summarization system to provide fluent yet precise summaries. In contrast to these works, we integrate a retrieval based method with a context-aware neural dialogue generation model in order to introduce relevant responses in a goal oriented conversation. 

\section{Conclusion}
In this study, we have experimented with a simple yet effective way of conditioning the decoder in a dialogue generation model intended for goal oriented dialogue. Generating fluent \textit{and} precise responses is crucial for creating goal-oriented dialogue systems, however, this can be a very difficult task; particularly, when the system responses are dependent on domain-specific information. We propose adding a simple retrieval step, where we obtain the past conversations that are most relevant to the current one and condition our decoder on these. We find that this method not only improves over multiple strong baseline models on word overlap metrics,  it also performs better than the state-of-the-art on BLEU and achieves competitive performance for inform/request metrics without requiring dialog act annotations. Finally, by inspecting the output of the baseline versus our proposed model and through human evaluations, we find that a great advantage of our model is its ability to produce responses that are more fluent and remain on topic.
\bibliography{neurips_2019}

\begin{thebibliography}{50}
\expandafter\ifx\csname natexlab\endcsname\relax\def\natexlab#1{#1}\fi

\bibitem[{Budzianowski et~al.(2018)Budzianowski, Wen, Tseng, Casanueva, Ultes,
  Ramadan, and Gasic}]{budzianowski2018multiwoz}
Pawe{\l} Budzianowski, Tsung-Hsien Wen, Bo-Hsiang Tseng, I{\~n}igo Casanueva,
  Stefan Ultes, Osman Ramadan, and Milica Gasic. 2018.
\newblock Multiwoz-a large-scale multi-domain wizard-of-oz dataset for
  task-oriented dialogue modelling.
\newblock In \emph{Proceedings of the 2018 Conference on Empirical Methods in
  Natural Language Processing}, pages 5016--5026.

\bibitem[{Chen et~al.(2017)Chen, Liu, Yin, and Tang}]{chen2017survey}
Hongshen Chen, Xiaorui Liu, Dawei Yin, and Jiliang Tang. 2017.
\newblock A survey on dialogue systems: Recent advances and new frontiers.
\newblock \emph{ACM SIGKDD Explorations Newsletter}, 19(2):25--35.

\bibitem[{Chen et~al.(2019)Chen, Chen, Qin, Yan, and
  Wang}]{chen2019semantically}
Wenhu Chen, Jianshu Chen, Pengda Qin, Xifeng Yan, and William~Yang Wang. 2019.
\newblock Semantically conditioned dialog response generation via hierarchical
  disentangled self-attention.
\newblock \emph{arXiv preprint arXiv:1905.12866}.

\bibitem[{Cho et~al.(2014)Cho, Gulcehre, Bahdanau, Schwenk, and
  Bengio}]{cholearning}
Kyunghyun Cho, Bart van Merri{\"e}nboer~Caglar Gulcehre, Dzmitry Bahdanau,
  Fethi Bougares~Holger Schwenk, and Yoshua Bengio. 2014.
\newblock Learning phrase representations using rnn encoder--decoder for
  statistical machine translation.

\bibitem[{Galley et~al.(2015)Galley, Brockett, Sordoni, Ji, Auli, Quirk,
  Mitchell, Gao, and Dolan}]{galley2015deltableu}
Michel Galley, Chris Brockett, Alessandro Sordoni, Yangfeng Ji, Michael Auli,
  Chris Quirk, Margaret Mitchell, Jianfeng Gao, and Bill Dolan. 2015.
\newblock deltableu: A discriminative metric for generation tasks with
  intrinsically diverse targets.
\newblock \emph{arXiv preprint arXiv:1506.06863}.

\bibitem[{Gonzalez et~al.(2018)Gonzalez, Augenstein, and
  S{\o}gaard}]{gonzalez2018strong}
Ana Gonzalez, Isabelle Augenstein, and Anders S{\o}gaard. 2018.
\newblock A strong baseline for question relevancy ranking.
\newblock In \emph{Proceedings of the 2018 Conference on Empirical Methods in
  Natural Language Processing}, pages 4810--4815.

\bibitem[{Henderson(2015)}]{henderson2015machine}
Matthew Henderson. 2015.
\newblock \href
  {https://static.googleusercontent.com/media/research.google.com/en//pubs/archive/44018.pdf}
  {Machine learning for dialog state tracking: A review}.
\newblock In \emph{Proc. of The First International Workshop on Machine
  Learning in Spoken Language Processing}.

\bibitem[{Henderson et~al.(2013)Henderson, Thomson, and
  Young}]{henderson2013deep}
Matthew Henderson, Blaise Thomson, and Steve Young. 2013.
\newblock Deep neural network approach for the dialog state tracking challenge.
\newblock In \emph{Proceedings of the SIGDIAL 2013 Conference}, pages 467--471.

\bibitem[{Henderson et~al.(2014)Henderson, Thomson, and
  Young}]{henderson2014word}
Matthew Henderson, Blaise Thomson, and Steve Young. 2014.
\newblock \href {https://www.aclweb.org/anthology/W/W14/W14-4340.pdf}
  {Word-based dialog state tracking with recurrent neural networks}.
\newblock In \emph{Proceedings of the 15th Annual Meeting of the Special
  Interest Group on Discourse and Dialogue (SIGDIAL)}, pages 292--299.

\bibitem[{Indyk and Motwani(1998)}]{Indyk:1998:ANN:276698.276876}
Piotr Indyk and Rajeev Motwani. 1998.
\newblock \href {https://doi.org/10.1145/276698.276876} {Approximate nearest
  neighbors: Towards removing the curse of dimensionality}.
\newblock In \emph{Proceedings of the Thirtieth Annual ACM Symposium on Theory
  of Computing}, STOC '98, pages 604--613, New York, NY, USA. ACM.

\bibitem[{Kingma and Ba(2014)}]{KingmaB14}
Diederik~P. Kingma and Jimmy Ba. 2014.
\newblock \href {http://arxiv.org/abs/1412.6980} {Adam: {A} method for
  stochastic optimization}.
\newblock \emph{ICLR}.

\bibitem[{Li et~al.(2015)Li, Luong, and Jurafsky}]{li2015hierarchical}
Jiwei Li, Thang Luong, and Dan Jurafsky. 2015.
\newblock A hierarchical neural autoencoder for paragraphs and documents.
\newblock In \emph{Proceedings of the 53rd Annual Meeting of the Association
  for Computational Linguistics and the 7th International Joint Conference on
  Natural Language Processing (Volume 1: Long Papers)}, volume~1, pages
  1106--1115.

\bibitem[{Li et~al.(2017{\natexlab{a}})Li, Monroe, Shi, Jean, Ritter, and
  Jurafsky}]{li2017adversarial}
Jiwei Li, Will Monroe, Tianlin Shi, S{\.e}bastien Jean, Alan Ritter, and Dan
  Jurafsky. 2017{\natexlab{a}}.
\newblock Adversarial learning for neural dialogue generation.
\newblock In \emph{Proceedings of the 2017 Conference on Empirical Methods in
  Natural Language Processing}, pages 2157--2169.

\bibitem[{Li et~al.(2017{\natexlab{b}})Li, Chen, Li, Gao, and
  Celikyilmaz}]{Li:ea:17}
Xiujun Li, Yun-Nung Chen, Lihong Li, Jianfeng Gao, and Asli Celikyilmaz.
  2017{\natexlab{b}}.
\newblock End-to-end task-completion neural dialogue systems.
\newblock In \emph{IJCNLP}.

\bibitem[{Lison and Tiedemann(2016)}]{lison2016opensubtitles2016}
Pierre Lison and J{\"o}rg Tiedemann. 2016.
\newblock Opensubtitles2016: Extracting large parallel corpora from movie and
  tv subtitles.

\bibitem[{Liu et~al.(2016)Liu, Lowe, Serban, Noseworthy, Charlin, and
  Pineau}]{liu2016not}
Chia-Wei Liu, Ryan Lowe, Iulian Serban, Mike Noseworthy, Laurent Charlin, and
  Joelle Pineau. 2016.
\newblock How not to evaluate your dialogue system: An empirical study of
  unsupervised evaluation metrics for dialogue response generation.
\newblock In \emph{Proceedings of the 2016 Conference on Empirical Methods in
  Natural Language Processing}, pages 2122--2132.

\bibitem[{Lowe et~al.(2017)Lowe, Noseworthy, Serban, A.-Gontier, Bengio, and
  Pineau}]{Lowe:ea:17}
Ryan Lowe, Michael Noseworthy, Iulian~V. Serban, Nicolas A.-Gontier, Yoshua
  Bengio, and Joelle Pineau. 2017.
\newblock Towards an automatic turing test: Learning to evaluate dialogue
  responses.
\newblock In \emph{ACL}.

\bibitem[{Lowe et~al.(2015)Lowe, Pow, Serban, and Pineau}]{lowe2015ubuntu}
Ryan Lowe, Nissan Pow, Iulian Serban, and Joelle Pineau. 2015.
\newblock The ubuntu dialogue corpus: A large dataset for research in
  unstructured multi-turn dialogue systems.
\newblock \emph{arXiv preprint arXiv:1506.08909}.

\bibitem[{Luan et~al.(2016)Luan, Ji, and Ostendorf}]{luan2016lstm}
Yi~Luan, Yangfeng Ji, and Mari Ostendorf. 2016.
\newblock Lstm based conversation models.
\newblock \emph{arXiv preprint arXiv:1603.09457}.

\bibitem[{Mrk{\v{s}}i{\'c} et~al.(2015)Mrk{\v{s}}i{\'c}, S{\'e}aghdha, Thomson,
  Ga{\v{s}}i{\'c}, Su, Vandyke, Wen, and Young}]{mrkvsic2015multi}
N~Mrk{\v{s}}i{\'c}, DO~S{\'e}aghdha, B~Thomson, M~Ga{\v{s}}i{\'c}, PH~Su,
  D~Vandyke, TH~Wen, and S~Young. 2015.
\newblock \href {https://aclanthology.info/papers/P15-2130/p15-2130}
  {Multi-domain dialog state tracking using recurrent neural networks}.
\newblock In \emph{ACL-IJCNLP 2015-53rd Annual Meeting of the Association for
  Computational Linguistics and the 7th International Joint Conference on
  Natural Language Processing of the Asian Federation of Natural Language
  Processing, Proceedings of the Conference}, volume~2, pages 794--799.

\bibitem[{Mrk{\v{s}}i{\'c} et~al.(2016)Mrk{\v{s}}i{\'c}, S{\'e}aghdha, Wen,
  Thomson, and Young}]{mrkvsic2016neural}
Nikola Mrk{\v{s}}i{\'c}, Diarmuid~O S{\'e}aghdha, Tsung-Hsien Wen, Blaise
  Thomson, and Steve Young. 2016.
\newblock Neural belief tracker: Data-driven dialogue state tracking.
\newblock \emph{arXiv preprint arXiv:1606.03777}.

\bibitem[{Mrk{\v{s}}i{\'c} et~al.(2017)Mrk{\v{s}}i{\'c}, S{\'e}aghdha, Wen,
  Thomson, and Young}]{mrkvsic2017neural}
Nikola Mrk{\v{s}}i{\'c}, Diarmuid~{\'O} S{\'e}aghdha, Tsung-Hsien Wen, Blaise
  Thomson, and Steve Young. 2017.
\newblock \href {http://aclweb.org/anthology/P17-1163} {Neural belief tracker:
  Data-driven dialogue state tracking}.
\newblock In \emph{Proceedings of the 55th Annual Meeting of the Association
  for Computational Linguistics (Volume 1: Long Papers)}, volume~1, pages
  1777--1788.

\bibitem[{Papineni et~al.(2002)Papineni, Roukos, Ward, and jing
  Zhu}]{Papineni02bleu:a}
Kishore Papineni, Salim Roukos, Todd Ward, and Wei jing Zhu. 2002.
\newblock Bleu: a method for automatic evaluation of machine translation.
\newblock pages 311--318.

\bibitem[{Rastogi et~al.(2017)Rastogi, Hakkani-T{\"u}r, and
  Heck}]{rastogi2017scalable}
Abhinav Rastogi, Dilek Hakkani-T{\"u}r, and Larry Heck. 2017.
\newblock \href
  {https://static.googleusercontent.com/media/research.google.com/en//pubs/archive/46399.pdf}
  {Scalable multi-domain dialogue state tracking}.
\newblock In \emph{2017 IEEE Automatic Speech Recognition and Understanding
  Workshop (ASRU)}, pages 561--568. IEEE.

\bibitem[{Rastogi et~al.(2019)Rastogi, Gupta, Chen, and
  Mathias}]{rastogi2019scaling}
Pushpendre Rastogi, Arpit Gupta, Tongfei Chen, and Lambert Mathias. 2019.
\newblock Scaling multi-domain dialogue state tracking via query reformulation.
\newblock \emph{arXiv preprint arXiv:1903.05164}.

\bibitem[{Ren et~al.(2013)Ren, Xu, Zhang, and Yan}]{ren2013dialog}
Hang Ren, Weiqun Xu, Yan Zhang, and Yonghong Yan. 2013.
\newblock Dialog state tracking using conditional random fields.
\newblock In \emph{Proceedings of the SIGDIAL 2013 Conference}, pages 457--461.

\bibitem[{Ren et~al.(2018)Ren, Xie, Chen, and Yu}]{ren2018towards}
Liliang Ren, Kaige Xie, Lu~Chen, and Kai Yu. 2018.
\newblock Towards universal dialogue state tracking.
\newblock In \emph{Proceedings of the 2018 Conference on Empirical Methods in
  Natural Language Processing}, pages 2780--2786.

\bibitem[{Ritter et~al.(2011)Ritter, Cherry, and Dolan}]{ritter2011data}
Alan Ritter, Colin Cherry, and William~B Dolan. 2011.
\newblock Data-driven response generation in social media.
\newblock In \emph{Proceedings of the conference on empirical methods in
  natural language processing}, pages 583--593. Association for Computational
  Linguistics.

\bibitem[{Serban et~al.(2015)Serban, Sordoni, Bengio, Courville, and
  Pineau}]{serban2015hierarchical}
Iulian~Vlad Serban, Alessandro Sordoni, Yoshua Bengio, Aaron~C Courville, and
  Joelle Pineau. 2015.
\newblock Hierarchical neural network generative models for movie dialogues.
\newblock \emph{CoRR, abs/1507.04808}.

\bibitem[{Serban et~al.(2016)Serban, Sordoni, Bengio, Courville, and
  Pineau}]{serban2016building}
Iulian~Vlad Serban, Alessandro Sordoni, Yoshua Bengio, Aaron~C Courville, and
  Joelle Pineau. 2016.
\newblock Building end-to-end dialogue systems using generative hierarchical
  neural network models.
\newblock In \emph{AAAI}, volume~16, pages 3776--3784.

\bibitem[{Serban et~al.(2017)Serban, Sordoni, Lowe, Charlin, Pineau, Courville,
  and Bengio}]{serban2017hierarchical}
Iulian~Vlad Serban, Alessandro Sordoni, Ryan Lowe, Laurent Charlin, Joelle
  Pineau, Aaron~C Courville, and Yoshua Bengio. 2017.
\newblock A hierarchical latent variable encoder-decoder model for generating
  dialogues.
\newblock In \emph{AAAI}, pages 3295--3301.

\bibitem[{Shang et~al.(2015{\natexlab{a}})Shang, Lu, and
  Li}]{shang2015responding}
Lifeng Shang, Zhengdong Lu, and Hang Li. 2015{\natexlab{a}}.
\newblock Neural responding machine for short-text conversation.
\newblock In \emph{ACL}.

\bibitem[{Shang et~al.(2015{\natexlab{b}})Shang, Lu, and Li}]{shang2015neural}
Lifeng Shang, Zhengdong Lu, and Hang Li. 2015{\natexlab{b}}.
\newblock Neural responding machine for short-text conversation.
\newblock In \emph{Proceedings of the 53rd Annual Meeting of the Association
  for Computational Linguistics and the 7th International Joint Conference on
  Natural Language Processing (Volume 1: Long Papers)}, volume~1, pages
  1577--1586.

\bibitem[{Shen et~al.(2017)Shen, Su, Li, Li, Niu, Zhao, Aizawa, and
  Long}]{shen2017conditional}
Xiaoyu Shen, Hui Su, Yanran Li, Wenjie Li, Shuzi Niu, Yang Zhao, Akiko Aizawa,
  and Guoping Long. 2017.
\newblock A conditional variational framework for dialog generation.
\newblock In \emph{Proceedings of the 55th Annual Meeting of the Association
  for Computational Linguistics (Volume 2: Short Papers)}, volume~2, pages
  504--509.

\bibitem[{Shibata et~al.(2009)Shibata, Nishiguchi, and
  Tomiura}]{shibata2009dialog}
Masahiro Shibata, Tomomi Nishiguchi, and Yoichi Tomiura. 2009.
\newblock Dialog system for open-ended conversation using web documents.
\newblock \emph{Informatica}, 33(3).

\bibitem[{Song et~al.(2016)Song, Yan, Li, Zhao, and Zhang}]{Song2016}
Yiping Song, Rui Yan, Xiang Li, Dongyan Zhao, and Ming Zhang. 2016.
\newblock Two are better than one: An ensemble of retrieval- and
  generation-based dialog systems.
\newblock \emph{arXiv preprint arXiv:1610.07149}.

\bibitem[{Sordoni et~al.(2015)Sordoni, Bengio, Vahabi, Lioma, Grue~Simonsen,
  and Nie}]{sordoni2015hierarchical}
Alessandro Sordoni, Yoshua Bengio, Hossein Vahabi, Christina Lioma, Jakob
  Grue~Simonsen, and Jian-Yun Nie. 2015.
\newblock A hierarchical recurrent encoder-decoder for generative context-aware
  query suggestion.
\newblock In \emph{Proceedings of the 24th ACM International on Conference on
  Information and Knowledge Management}, pages 553--562. ACM.

\bibitem[{Subramanian et~al.(2019)Subramanian, Li, Pilault, and
  Pal}]{subramanian2019extractive}
Sandeep Subramanian, Raymond Li, Jonathan Pilault, and Christopher Pal. 2019.
\newblock On extractive and abstractive neural document summarization with
  transformer language models.
\newblock \emph{arXiv preprint arXiv:1909.03186}.

\bibitem[{Sugiyama et~al.(2013)Sugiyama, Meguro, Higashinaka, and
  Minami}]{sugiyama2013open}
Hiroaki Sugiyama, Toyomi Meguro, Ryuichiro Higashinaka, and Yasuhiro Minami.
  2013.
\newblock Open-domain utterance generation for conversational dialogue systems
  using web-scale dependency structures.
\newblock In \emph{Proceedings of the SIGDIAL 2013 Conference}, pages 334--338.

\bibitem[{Sun et~al.(2014)Sun, Chen, Zhu, and Yu}]{sun2014generalized}
Kai Sun, Lu~Chen, Su~Zhu, and Kai Yu. 2014.
\newblock A generalized rule based tracker for dialogue state tracking.
\newblock In \emph{2014 IEEE Spoken Language Technology Workshop (SLT)}, pages
  330--335. IEEE.

\bibitem[{Sun et~al.(2016)Sun, Zhu, Chen, Yao, Wu, and Yu}]{sun2016hybrid}
Kai Sun, Su~Zhu, Lu~Chen, Siqiu Yao, Xueyang Wu, and Kai Yu. 2016.
\newblock Hybrid dialogue state tracking for real world human-to-human
  dialogues.
\newblock In \emph{INTERSPEECH}, pages 2060--2064.

\bibitem[{Vaswani et~al.(2017)Vaswani, Shazeer, Parmar, Uszkoreit, Jones,
  Gomez, Kaiser, and Polosukhin}]{vaswani2017attention}
Ashish Vaswani, Noam Shazeer, Niki Parmar, Jakob Uszkoreit, Llion Jones,
  Aidan~N Gomez, {\L}ukasz Kaiser, and Illia Polosukhin. 2017.
\newblock Attention is all you need.
\newblock In \emph{Advances in neural information processing systems}, pages
  5998--6008.

\bibitem[{Vinyals and Le(2015)}]{vinyals2015neural}
Oriol Vinyals and Quoc~V Le. 2015.
\newblock A neural conversational model.

\bibitem[{Wen et~al.(2018)Wen, Liu, Che, Qin, and Liu}]{wen2018kb}
Haoyang Wen, Yijia Liu, Wanxiang Che, Libo Qin, and Ting Liu. 2018.
\newblock Sequence-to-sequence learning for task-oriented dialogue with
  dialogue state representation.
\newblock In \emph{COLING}.

\bibitem[{Wen et~al.(2015)Wen, Gasic, Mrksic, Su, Vandyke, and
  Young}]{wen2015semantically}
Tsung-Hsien Wen, Milica Gasic, Nikola Mrksic, Pei-Hao Su, David Vandyke, and
  Steve Young. 2015.
\newblock Semantically conditioned lstm-based natural language generation for
  spoken dialogue systems.
\newblock \emph{arXiv preprint arXiv:1508.01745}.

\bibitem[{Weston et~al.(2018)Weston, Dinan, and Miller}]{weston2018retrieve}
Jason Weston, Emily Dinan, and Alexander Miller. 2018.
\newblock Retrieve and refine: Improved sequence generation models for
  dialogue.
\newblock In \emph{Proceedings of the 2018 EMNLP Workshop SCAI: The 2nd
  International Workshop on Search-Oriented Conversational AI}, pages 87--92.

\bibitem[{White and Caldwell(1998)}]{white1998exemplars}
Michael White and Ted Caldwell. 1998.
\newblock Exemplars: A practical, extensible framework for dynamic text
  generation.
\newblock \emph{Natural Language Generation}.

\bibitem[{Yoshino et~al.(2016)Yoshino, Hiraoka, Neubig, and
  Nakamura}]{yoshino2016dialogue}
Koichiro Yoshino, Takuya Hiraoka, Graham Neubig, and Satoshi Nakamura. 2016.
\newblock Dialogue state tracking using long short term memory neural networks.
\newblock In \emph{Proceedings of Seventh International Workshop on Spoken
  Dialog Systems}, pages 1--8.

\bibitem[{Zhang et~al.(2018)Zhang, Lan, Guo, Xu, and
  Cheng}]{zhang2018coherence}
Hainan Zhang, Yanyan Lan, Jiafeng Guo, Jun Xu, and Xueqi Cheng. 2018.
\newblock Reinforcing coherence for sequence to sequence model in dialogue
  generation.
\newblock In \emph{IJCAI}.

\bibitem[{Zhao and Eskenazi(2016)}]{zhao2016}
Tiancheng Zhao and Maxine Eskenazi. 2016.
\newblock \href {https://doi.org/10.18653/v1/W16-3601} {Towards end-to-end
  learning for dialog state tracking and management using deep reinforcement
  learning}.
\newblock pages 1--10.

\end{thebibliography}
\bibliographystyle{neurips_2019}

\end{document}